\documentclass[10pt,twocolumn,letterpaper]{article}

\usepackage{iccv}              % To produce the CAMERA-READY version
\usepackage{graphicx}
\usepackage{amsmath}
\usepackage{amssymb}
\usepackage{booktabs}

\usepackage{tabularx}
\usepackage{multirow}
\usepackage{multicol}
\usepackage{array}
\usepackage{setspace}
\usepackage{xcolor}  
\usepackage{wrapfig}
\usepackage{subcaption}

\usepackage{xspace}
\makeatletter
\DeclareRobustCommand\onedot{\futurelet\@let@token\@onedot}
\def\@onedot{\ifx\@let@token.\else.\null\fi\xspace}

\def\eg{\emph{e.g}\onedot} 
\def\ie{\emph{i.e}\onedot} 
 
\def\etc{\emph{etc}\onedot}

\makeatother

\definecolor{iccvblue}{rgb}{0.21,0.49,0.74}
\usepackage[pagebackref,breaklinks,colorlinks,allcolors=iccvblue]{hyperref}

\def\paperID{15620} % *** Enter the Paper ID here
\def\confName{ICCV}
\def\confYear{2025}

\title{An Attention-based Model for Robust Forecasting with Missing Modality}

\author{Zhitian Zhang\textsuperscript{1}\thanks{Work done during Internship at RBC Borealis.} \quad
Wenjie Zi\textsuperscript{2} \quad
Yunduz Rakhmangulova\textsuperscript{2} \quad
Saghar Irandoust\textsuperscript{2} \\
Hossein Hajimirsadeghi\textsuperscript{2} \quad
Thibaut Durand\textsuperscript{2} \\ \\
\textsuperscript{1}Simon Fraser University \quad
\textsuperscript{2}RBC Borealis
}

\begin{document}
\maketitle
\begin{abstract}
Learning with missing modalities is a fundamental challenge in multimodal robot learning, as real-world robotic systems often operate in environments with incomplete sensor data. Attention-based models are appealing for processing multimodal data because they can handle multiple modalities with a single backbone network. However, most multimodal models assume that all modalities are available during both training and inference, limiting their applicability in robotic perception and decision-making.
In this paper, we introduce a multimodal model designed to handle missing modalities during both training and inference. The model is formulated as a conditional variational autoencoder (CVAE) and incorporates a transformer-based architecture that leverages attention mechanisms to learn a unified, fixed-dimensional representation, even when some modalities are missing. We show that our proposed model can be trained with missing modalities while approximating a robust representation of all modalities.
We evaluate our approach on five multimodal datasets across two robot learning tasks: human trajectory prediction and robot manipulation forecasting. Experimental results demonstrate that our model effectively learns from incomplete data, and is superior to prior multimodal fusion approaches.
\end{abstract}
\section{Introduction}

Predicting the future is usually a hard problem that requires one to understand and reason over the surrounding environment,
% For example, self-driving cars need to understand the trajectory of the other agents (\eg other cars, pedestrians, bikes, \etc) to make safe decisions \todo{add ref}.
%Modeling and understanding the surrounding environment is a very challenging task because the environment is usually complex 
and its perception is inherently multimodal--we see objects, hear sounds, feel textures, smell odors, \etc \citep{Ghazanfar2006}.
% A lot of sensors have been developed to help machines to capture different parts of the environment. 
This is especially true in applications like autonomous driving \citep{nuscenes2020, sun2020scalability, girase2021loki} and video understanding \citep{sun2019videobert, gberta_2021_ICML, liu2021video} where it is necessary to comprehend and reason over several modalities to solve a problem.
% multimodal learning has become the dominant approach. 
For instance, autonomous navigation algorithms can incorporate multiple types of sensor data, such as LiDAR, camera, GPS, and odometer, to have a better understanding of the environment and make more informed decisions. 

\begin{figure}[t]
  \centering
\includegraphics[width=0.95\linewidth]{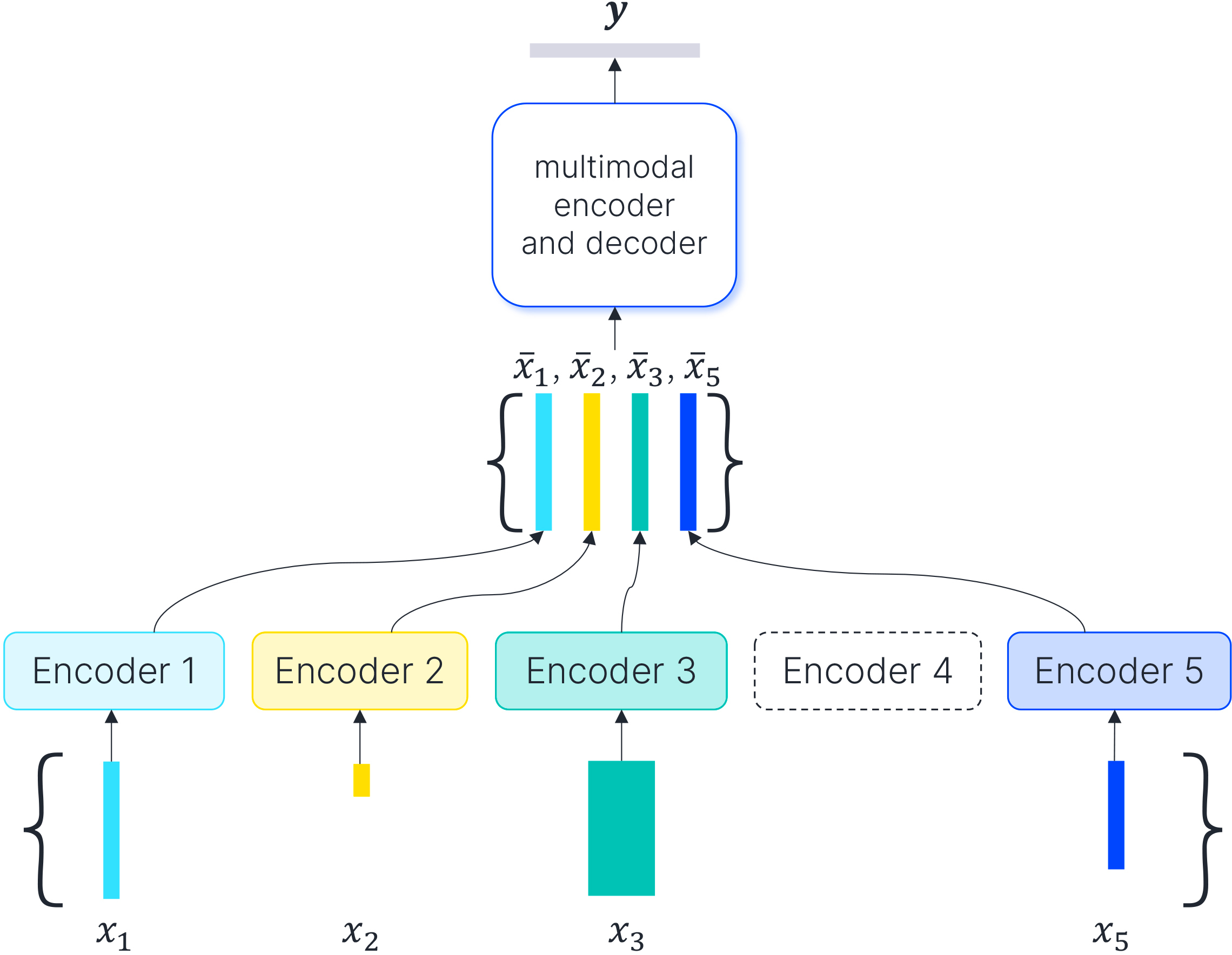}
\caption{Overall architecture of the proposed model with 5 modalities.
The model makes predictions based on the available modalities.
In this example, four modalities are available: 
$\{\mathbf{x}_1,\mathbf{x}_2,\mathbf{x}_3,\mathbf{x}_5\}$.
The 4th modality (${x}_4$) is missing and its associated encoder is represented with dotted lines.
The model can adapt to several types of inputs.
These inputs can have different dimensions or types (\eg vector and image).
Each unimodal encoder learns a fixed-dimension vectorial representation $\mathbf{\bar{x}}_i$ for each modality $\mathbf{{x}}_i$.
Then, an attention-based encoder-decoder model learns multimodal representations and makes the predictions.
}
\label{figure:octopus_architecture}
\end{figure}

In the past years, multimodal machine learning has become a dominating approach in multiple areas of computer vision.
Existing multimodal models \citep{Arandjelovic2018, sun2019videobert, xiao2020audiovisual, nagrani2021attention} require modal-complete data to achieve optimal performance.
%\ie they work only if all the examples have all the modalities during both training and inference. 
This constraint may be difficult to hold in real-world applications and limit the use cases.
Ideally, we would like to design models (\autoref{figure:octopus_architecture}) that can work on examples with missing modalities during both training and inference because they have several advantages.
First, during the training phase, it is possible to improve the model performance by training the model on a larger dataset. 
\cite{Sun2017,Mahajan2018} show that increasing the training dataset size can be one way to improve performance significantly.
Building a large multimodal dataset is expensive and time-consuming, 
%because it requires to collect the data for all modalities, clean them, and align them.
and it is easier and less costly to build a large multimodal dataset where some examples have missing modalities.
% \todo{this seems to be part of the Second advantage but not listed there?} 
%Examples with missing modalities should not be ignored during training because they may contain valuable information like rare examples.
Second, during the inference phase, the model will be robust to examples with missing modalities \ie the model can adapt to examples with missing modalities. 
% For example, if one of the sensors has an outage, which is a scenario quite frequent in real world applications, the model will not break and continue to make predictions. 
%Note that, for a given example, the quality of the prediction made with a missing modality may be lower than a prediction made with all the modalities.
% It is possible to build larger training datasets because it is not necessary to collect all the modalities of all the examples.
% We can build a dataset where some modalities of some examples are missing.
% We think that designing models that can handle examples with missing modalities can help to scale multimodal machine learning to larger datasets and have a large impact in the future.
% It can helps to build larger datasets. 
Previous works \citep{ma2021smil, pham2019found} on missing-modality learning mostly exist in the image-text domain where there are only two modalities. In the robotics domain, we often have data with more than two modalities, and such data are limited. Therefore, we are interested in developing novel methods that can learn a multimodal model from more than two modalities, and maintain a comparable performance when there is a missing modality.
In particular, we specifically focus on multimodal fusion applied in trajectory prediction tasks in this work. 
We first study on the human trajectory prediction task and then further extend our approach to the robot manipulation task. Multimodal fusion approaches have been extensively studied for both tasks. However, the missing-modality scenarios in these tasks are not properly addressed. We consider the scenario where three modalities are available during training and one of the modalities is entirely missing during inference.

We formulate our model in the Conditional Variational Auto-Encoder (CVAE) paradigm \citep{Kihyuk2015}. In addition to the evidence lower-bound (ELBO), we design a missing-modality loss for our CVAE. This loss allows our model to implicitly reconstruct the missing modality and generate representations as close as possible to the ones from full modalities. Our CVAE leverages the adaptive nature of attention mechanism \citep{lee2019set} as our modality aggregation scheme, and random masking technique to learn multimodal representations with available modalities. Specifically, our ``modality-agnostic attention'' aggregation scheme combines all available modalities to form a fixed dimensional latent, allowing our model to be robust to a different number of modalities.

To summarize, our contribution is threefold. 
\textbf{(1)} We propose a simple yet efficient modality aggregation scheme for the multimodal forecasting tasks using attentions, which is agnostic to missing modalities during both training and inference.
%model architecture for multimodal forecasting tasks, using attention as a unique modality aggregation scheme, that is agnostic to missing modalities during both \textbf{training} and \textbf{inference}.
%can handle multimodal examples with missing modalities during both \textbf{training} and \textbf{inference}, allowing our model being robust to missing modalities.
%Unlike most of the multimodal models that work only in a bimodal setting, our model can work on tasks with more than two modalities.
\textbf{(2)} We present a robust multimodal learning paradigm where a conditional probability distribution generated with missing modality is approximated to be as close as the one generated with full modalities.
\textbf{(3)} We extensively evaluate our model on human trajectory and robot manipulation forecasting tasks with missing modalities, and show that our proposed model outperforms existing baselines.
We analyze the performances of our model in multiple scenarios of missing modalities. 

% To summarize, our contribution is threefold. 
% \textbf{(1)} We propose a new model architecture, using attention, that can handle multimodal examples with missing modalities during both training and inference. 
% Unlike most of the multimodal models that work only in a bimodal setting, our model can work on tasks with more than two modalities.
% \textbf{(2)} We propose a specific instantiation of our model for the human trajectory prediction task.
% \textbf{(3)} We evaluate our model on human trajectory prediction datasets with missing modalities and show that our proposed model outperforms existing baselines.
% We analyze the performances of our model in multiple scenarios of missing modalities. 

% We extend the evaluation strategy to evaluate the performance of our model on multimodal trajectory prediction datasets with missing modalities because existing multimodal trajectory prediction benchmarks focus a modal-complete scenario. 
% We show that our proposed model outperforms existing baselines

\section{Related Work}

\paragraph{Multimodal Learning. }Multimodal learning seeks to improve the performance for a task by utilizing the different available modalities, such as images, sound and natural language.  
The importance and availability of these modalities are usually not uniform \cite{liang2022foundations, liang2021multibench, ma2022multimodal}.  
Therefore it is critical to have an efficient multimodal fusion method for multimodal learning. Early approaches for multimodal learning are simple methods like concatenation \cite{poria2016convolutional, wang2017select}, tensor fusion \cite{zadeh2017tensor, ben2017mutan}, pooling fusions \cite{ramachandram2017deep}. Surprisingly late fusion method \cite{liang2021multibench} shows promising results across various tasks and proven it is a modality-agnostic methods that balances complexity and performance. 
Later more deep learning methods are proposed to learn the multimodal representations. 
Generative adversarial networks (GANs) have been used in various approaches for multimodal representation learning \cite{isola2017image, lee2018diverse, zhu2017unpaired}.
CVAE \cite{NIPS2015_8d55a249} and conditional multimodal autoencoder \cite{pandey2017variational} are introduced, to learn cross-modal representations conditioned on all modalities. Multimodal VAE (MVAE) \cite{wu2018multimodal} instead learns a joint-modal representations of multimodal data by using the product-of-expert (POE) to learn the joint distribution of different modalities. The assumption for MVAE to approximate an accurate joint-modal representation is that each modality is conditionally independent. This kind of assumption is often not accurate when the multimodal inputs are not independent of each other. However, in our work, we do not make such an assumption.
%These methods require modal-complete data, for both training and testing, which is not robust to examples with missing modalities.
\vspace{-2mm}
\paragraph{Learning with missing modalities. } Based on the foundation of the MVAE \cite{wu2018multimodal}, SMIL \cite{ma2021smil} proposed to reconstruct the missing modalities from a modality prior even when the training modalities are missing. While some methods \cite{woo2023towards, zhao2021missing} like SMIL aimed to reconstruct the missing modalities, other approaches \cite{pham2019found, tsai2018learning, li2023multi, maheshwari2024missing} decided to learn a joint representation from all modalities. 
%Based on the foundation of the MVAE framework \cite{wu2018multimodal}, it was further extended to handle missing modality in the multimodal learnin scenario: SMIL \cite{ma2021smil} proposed to reconstruct the missing modalities from a modality prior even when the training modalities are missing. 
For example, MCTN \cite{pham2019found} was designed to learn a joint representation between a source modality and multiple target modalities during training, and infer the joint representation from only the source modality during testing. %which makes MCTN robust to missing information on modalities.
%\todo{we should maybe indicate the key result of the analysis}. 
%MMF \cite{tsai2018learning} learns a joint representation of multimodal data, and performs reconstruction of missing modalities during testing without significant impact on performance.
MissMod \cite{lin2023missmodal} proposed three constraints during learning to align the representations of missing and complete modalities for sentiment analysis. Similarly, \cite{garcia2019dmcl} trains individual modality networks and uses stronger modality networks to improve others.
And ActionMAE \cite{woo2023towards} learns a robust representation by randomly dropping modality features for action recognition. 
In this work, we propose a training scheme that randomly drops some modalities during training, while maximizing the alignment between the predicted outputs from complete and incomplete modalities.
%we propose an attention-based architecture that can automatically adapt to the number of available modalities.
%To train the model to be robust to missing modalities, we propose a training scheme that randomly drops some modalities during training, while maximizing the alignment with the predicted output without the randomly dropped modalities.
% However these approaches are either: (1) often limited to two modalities and do not extend easily to three or more modalities, or (2) do not work when modality is completely missing. Instead, we adopt our model to work with more than two modalities, and complete missing modality.
\vspace{-2mm}
\paragraph{Attentions.} Attention mechanism has been widely studied for multimodal fusion recently \cite{nagrani2021attention, sun2019videobert, ge2023metabev}. A number of multimodal transformers \cite{kim2021vilt, ye2019cross, strudel2021segmenter, tsai2019multimodal, woo2023towards} have been proposed in different domains.
\cite{ma2022multimodal} investigates the robustness of the transformer model on missing-modal data and discovers that the transformer model's performance drops dramatically with modal-complete data when processing all modalities together using self-attention mechanisms. Attention bottlenecks \cite{nagrani2021attention} proposed a ``fusion bottlenecks'' using cross-attention mechanism for multimodal fusion. 
%Standard positional encoding is used as bottleneck tokens which only allows useful information to flow through different modalities. 
The permutation-invariant property of the transformer model has also been studied in \cite{lee2019set, tang2021sensory} and applied on set-structured data which allows the network to work with an arbitrary number of inputs. 
However, the missing modality scenario was not part of their work. 
In this paper, we utilize an attention-based model to model multimodal inputs from both modal-complete and modal-incomplete data and test it with either full or missing modality. 

% \clearpage
% \newpage
\section{Background: Attention}

% \paragraph{Attention.}
Transformers \cite{vaswani2017attention} are built upon attention, which enables to capture relationships for tokens at different positions.
The attention receives three input sequences, namely
query $\mathbf{Q} \in \mathbb R^{n_q \times d}$, key $\mathbf{K} \in \mathbb R^{n_k \times d}$ and value $\mathbf{V} \in \mathbb R^{n_v \times d}$ where $n_q, n_k$ and $n_v$ are the sequence lengths of query, key and value respectively.
\begin{align}
\text{Att}(\mathbf{Q}, \mathbf{K}, \!\mathbf{V}) \!=\! \text{softmax} \left( \frac{\mathbf{Q} \mathbf{W}_Q (\mathbf{K} \mathbf{W}_K)^T}{\sqrt{d_h}} \right) \!\mathbf{V} \mathbf{W}_v
\end{align}
where $\mathbf{W}_q, \mathbf{W}_k, \mathbf{W}_v \!\in\! \mathbb R^{d \times d_h}$ are learnable parameters and $d_h$ is the number of hidden dimensions.
The self-attention (SA) is a special case of attention where $\mathbf{Q} = \mathbf{K} = \mathbf{V}$:
$\text{SA}(\mathbf{X}) = \text{Att}(\mathbf{X}, \mathbf{X}, \mathbf{X})$.
Multihead self-attention (MSA) is an extension of SA in which we run $k$ self-attention in parallel, and project their concatenated outputs:
% To keep compute and number of parameters constant when changing $k$, $d$ is typically set to $d/k$.
\begin{align}
\text{MSA}(\mathbf{X}) = [\text{SA}_1(\mathbf{X}), \text{SA}_2(\mathbf{X}), \ldots, \text{SA}_k(\mathbf{X})] \mathbf{W}_o
\end{align}
where $\mathbf{W}_o$ are learnable parameters.
Similarly, the multihead attention is:
\begin{align}
\text{MHA}(\!\mathbf{Q}, \!\mathbf{K}, \!\!\mathbf{V}\!) \!=\! [\text{Att}_1(\!\mathbf{Q}, \!\mathbf{K}, \!\!\mathbf{V}\!), \ldots, \text{Att}_k(\!\mathbf{Q}, \!\mathbf{K}, \!\!\mathbf{V}\!)] \mathbf{W}_o \!\!
\end{align}

\section{Proposed Approach}

\begin{figure*}[t]
\begin{center}
%\fbox{\rule{0pt}{2in} \rule{.9\linewidth}{0pt}}
\includegraphics[width=0.9\textwidth]{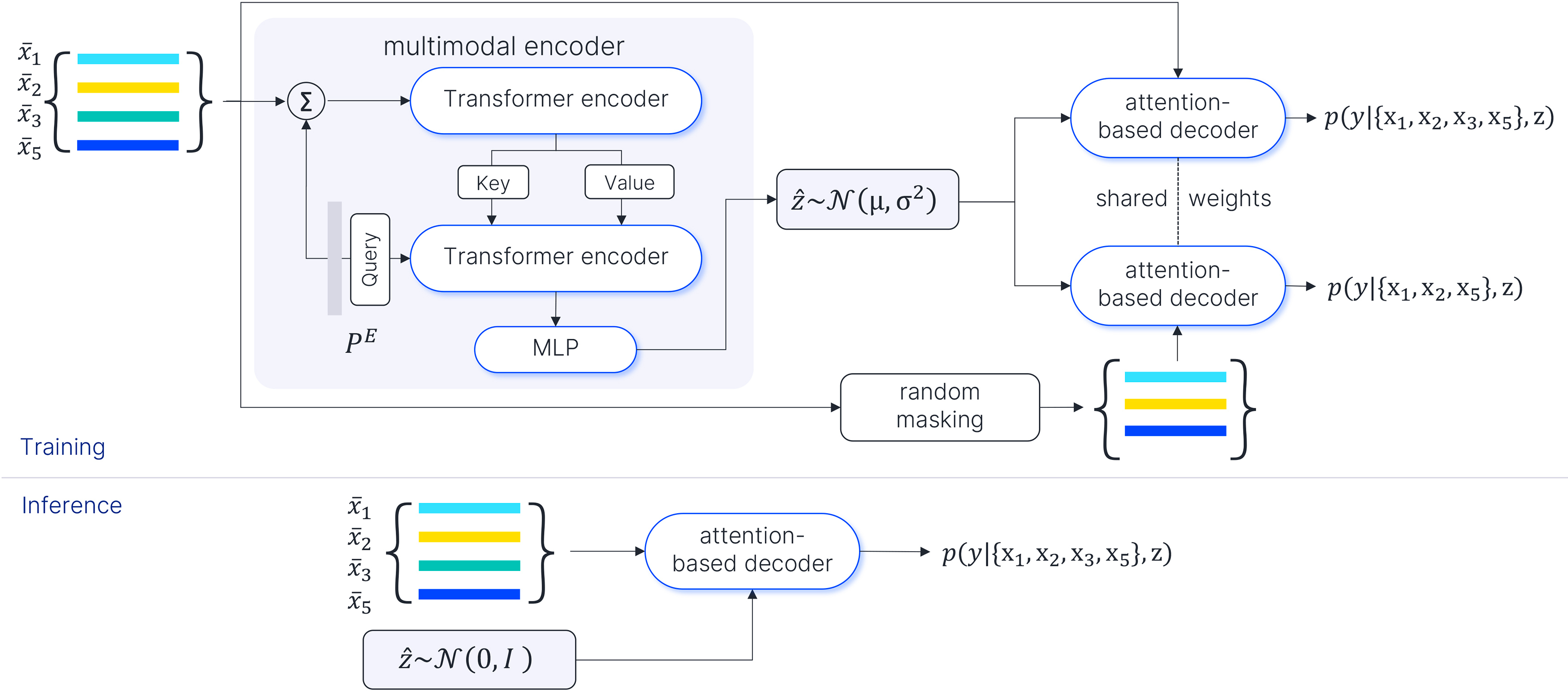}
\end{center}
\caption{Overview of our model during training (top) and inference (bottom).
The input to the multimodal encoder is a set of available modality representations obtained from the unimodal encoders (not represented in the figure). 
The multimodal encoder learns to generate a distribution over latent codes.
Then, a sampled latent code is decoded into a conditional probability distribution.
To train the model to be robust to missing modalities, the sampled latent code is decoded two times, with two sets of modalities: $\{\mathbf{x}_1,\mathbf{x}_2,\mathbf{x}_3,\mathbf{x}_5\}$ and $\{\mathbf{x}_1,\mathbf{x}_2,\mathbf{x}_5\}$.
This example shows $\mathbf{x}_3$ is removed but it can be any modalities. 
$\Sigma$ indicates the summation operator.
}
\label{fig:overall}
\end{figure*}

In this paper, we introduce a prediction model that can handle missing modalities during both training and inference without retraining or any modification to the model.
% To model the uncertainity of predicting the future, we use a CVAE.
First, we define some notations and introduce the model.
Then, we explain the training process.
Finally, we describe the model architecture.

\vspace{-2mm}
\paragraph{Notations.}
We denote by $M \geq 1$ the number of modalities and $\mathbb M = \{1, \ldots, M\}$ the set of all the modalities.
Let $\mathbb M^{(i)} \subseteq \mathbb M$ be the set of available modalities associated with the $i$-th example, and $M^{(i)}$ is the number of available modalities.
We assume $\mathbb M^{(i)}$ cannot be empty, so each example has at least one modality.
We denote the training data by $\mathcal{D} = \{(\mathbb X^{(1)}, \mathbf{y}^{(1)}), \ldots, (\mathbb X^{(N)}, \mathbf{y}^{(N)})\}$ where $\mathbb X^{(i)} = \{\mathbf{x}_m^{(i)}\}_{m \in \mathbb M^{(i)}}$ is the multimodal representations associated to the $i$-th example, and $\mathbf{y}^{(i)}$ is its ground-truth label.
$\mathbf{x}_m^{(i)}$ is the unimodal representation of the $m$-th modality of the $i$-th example.
The nature of $\mathbf{x}_m^{(i)}$ depends on the modality and each modality can be different \ie the first modality can be an image, the second one can be a time series and the third one can be a human pose. %, so $\mathbf{x}_m^{(i)}$ can be a vector or a matrix or a tensor.
% For example if the $m$-th modality is an RGB image, $\mathbf{x}_m^{(i)} \in \mathbb R^{H \times W \times 3}$ is a 3D tensor where $H$ and $W$ are the height and width of the image.
We are focusing on future prediction $\mathbf{y}^{(i)} = (\mathbf{y}^{(i)}_1, \ldots, \mathbf{y}^{(i)}_T)$ where $T$ is the number of future time steps to predict and $\mathbf{y}^{(i)}_t$ is the prediction at time step $t$.
In this paper, we assume the variable to predict is unimodal.
The history of the variable to predict can be one of the input modalities of the model.
% This formulation also supports non-temporal predictions by formulating the problem as a temporal problem with a single time step $T=1$.
% \tibo{The model is formulated as a temporal model to be generic, but it can be used on non-temporal problems by formulating the problem as a temporal problem with a single time step.}
% \todo{unimodal. Note that variable to predict can be part of the input}

% \paragraph{}
% We denote by $N$ the number of training examples and $M$ the number of modalities.
% We denote the training data by $\mathcal{D} = \{(\mathbb X^{(1)}, \mathbf{y}^{(1)}), \ldots, (\mathbb X^{(N)}, \mathbf{y}^{(N)})\}$ where $\mathbb X^{(i)} = \{\mathbf{x}_m^{(i)}\}_{m \in }$ is the $i$-th multimodal example.
% $\mathbb M^{(i)} \subset \mathbb M$
%\vspace{-6mm}
\vspace{-2mm}
\paragraph{Model overview.}
We formulate the future prediction with missing modalities problem as a generative model for a set of modalities using a Conditional Variational Auto-Encoder (CVAE) framework \cite{Kihyuk2015}.
We use a VAE formulation because we aim to develop a probabilistic model to capture the uncertainty in predicting the future.
During the training, the goal is to learn a function that outputs the probability distribution over the output $\mathbf{y}$ conditional a set of available modalities $\mathbb X$ representing the past $p(\mathbf{y}|\mathbb X; \Theta)$, where $\Theta$ are the whole set of model parameters.

\subsection{Model Architecture}
The overall architecture is shown in \autoref{fig:overall}, and has three main components: the unimodal encoders, the multimodal encoder, and the decoder.
First, the unimodal encoders extract fixed-dimensional vectorial representation for each available modality.
Then, the multimodal encoder learns a joint representation and outputs the parameters of the posterior distribution.
Finally, an attention-based decoder takes as input a latent code sampled from the posterior distribution and the set of available modalities to generate the output distribution.

\vspace{-2mm}
\paragraph{Unimodal encoders.}
We denote by $E_m$ the unimodal encoder for the $m$-th modality, and $d$ the dimension of the common space.
$\mathbf{\bar{x}}^{(i)}_m \in \mathbb R^{d}$ is the fixed-dimensional vectorial representation of the $m$-th modality of the $i$-th example.
In this work, we assume each modality is encoded separately but it is possible to extend the model to jointly encode several modalities.
The output of the unimodal encoders is a set of fixed dimensional vectorial representations:
\begin{align}
\bar{\mathbb X}^{(i)} = \left\{\mathbf{\bar{x}}^{(i)}_m\right\}_{m \in \mathbb M^{(i)}} \,\,\,\, \text{with} \,\,\,\,
\mathbf{\bar{x}}^{(i)}_m = E_m(\mathbf{x}^{(i)}_m) \in \mathbb R^{d}
\end{align}
The architecture of each unimodal encoder depends on the structure of the modality.
For example, a ConvNet can be used to encode an RGB image, and a GRU \citep{Chung2014} can be used to encode a sequence of vectors.

\vspace{-2mm}
\paragraph{Fusion via modality-agnostic attention.}
Our multimodal encoder model consists of two layers of the Transformer encoder \citep{vaswani2017attention} with some modifications. 
The first layer learns features across the available modalities (with or without missing modality), and the second layer leverages our modality aggregation scheme to produce a latent distribution that is agnostic to missing modalities. Inspired by \citep{lee2019set}, we proposed to aggregate multimodal features with missing modality by applying attention on a fixed-size learnable embedding. 
First, the available modalities are represented as a sequence of $M^{(i)}$ vectors by using the output of the unimodal encoders.
Position embeddings \citep{vaswani2017attention} $\mathbf{P}^E = [\mathbf{p}^E_1, \ldots, \mathbf{p}^E_M] \in \mathbb  R^{M \times d}$ are added to the sequence to retain modality information.
The resulting sequence of vectors serves as input to a Transformer encoder of $L$ layers.
The Transformer encoder \citep{vaswani2017attention} consists of alternating layers of multihead self-attention (MSA) and Multilayer Perceptron (MLP).
LayerNorm (LN) is applied before every block and residual connections after every block.
The MLP contains two layers with a ReLU non-linearity.
\begin{align}
\mathbf X_0^{(i)} \!&=\! \left[\mathbf{\bar{x}}^{(i)}_m + \mathbf{p}^E_m\right]_{m \in \mathbb M^{(i)}} \in \mathbb  R^{M^{(i)} \times d}
\\
\mathbf{X'}_l^{(i)} \!&=\! \text{MSA}(\text{LN}(\mathbf{X}_{l-1}^{(i)})) + \mathbf{X}_{l-1}^{(i)} \quad l \!=\! 1, \ldots, L\!-\!1
\\
\mathbf{X}_l^{(i)} \!&=\! \text{MLP}(\text{LN}(\mathbf{X'}_{l}^{(i)})) + \mathbf{X'}_{l}^{(i)} \,\,\quad l \!=\! 1, \ldots, L\!-\!1
\end{align}
The second layer of our multimodal encoder is designed to handle missing modalities during training time. We aggregate features over missing modalities by applying multihead attention on a learnable fixed-size embedding. We make query \textbf{Q} in our MHA operation to always have a fixed dimension as full modalities input so that the change in modality size does not affect the output size. Here we choose $\mathbf{P}^E$ as our query \textbf{Q}.
\begin{align}
\mathbf{X'}_L^{(i)} \!&=\! \text{MHA}(\text{LN}(\mathbf{P}^E), \text{LN}(\mathbf{X}_{L\!-\!1}^{(i)}), \text{LN}(\mathbf{X}_{L\!-\!1}^{(i)})) \!+\! \mathbf{X}_{L\!-\!1}^{(i)}
\\
\mathbf{X}_L^{(i)} \!&=\! \text{MLP}(\text{LN}(\mathbf{X'}_{L}^{(i)})) + \mathbf{X'}_{L}^{(i)}
\end{align}
The output $\mathbf{X}_L^{(i)}$ is flatten and passed into a MLP $f_\phi$ that outputs the posterior Gaussian distribution parameters:
\begin{align}
\mu_\phi^{(i)}, \sigma_\phi^{(i)} = f_\phi(\text{flat}(\mathbf{X}_L^{(i)}))
\end{align}

\begin{figure}[t]
\begin{center}
\includegraphics[width=0.99\linewidth]{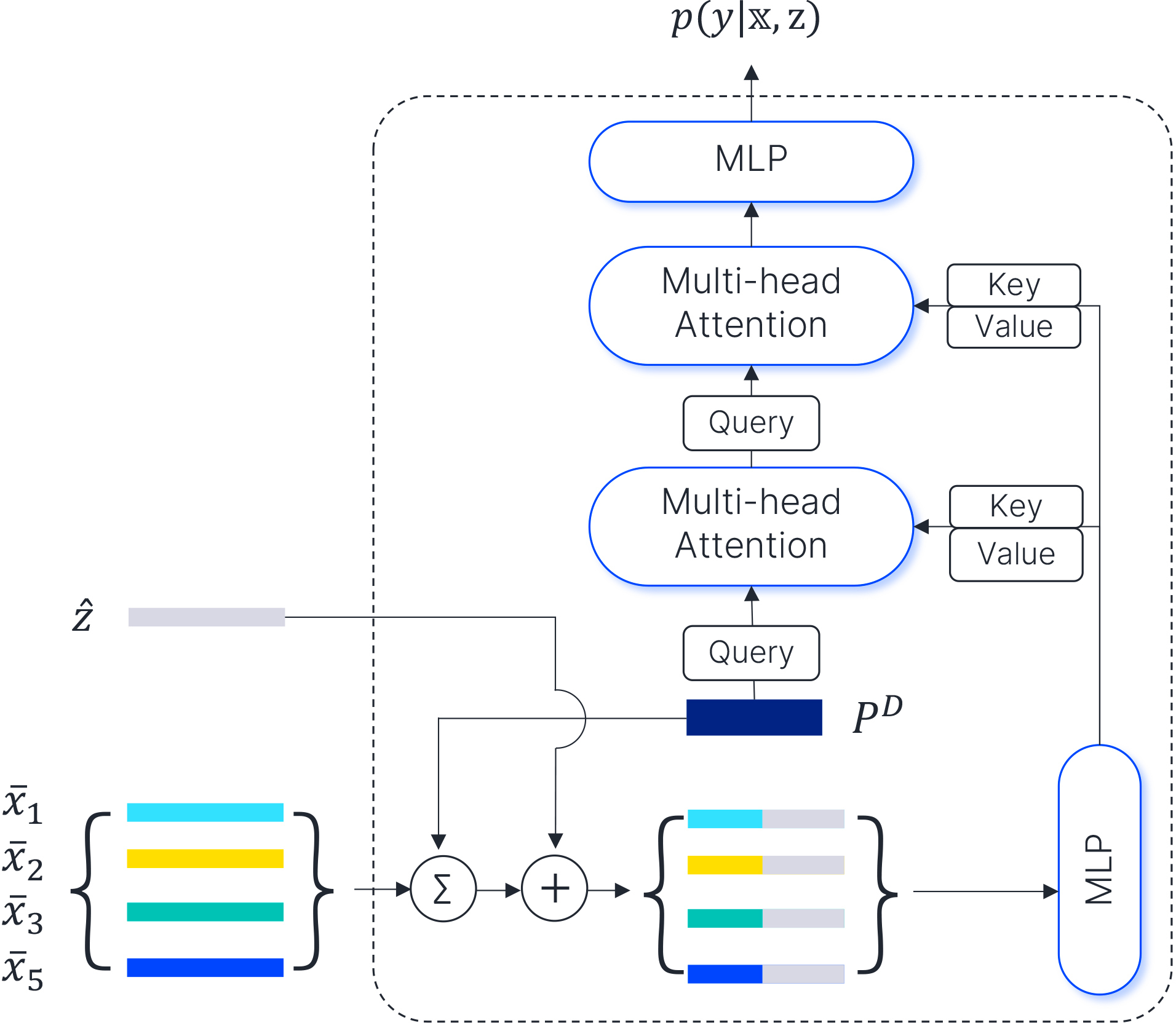}
\end{center}
\caption{Architecture of the attention-based decoder. The decoder takes the set of available modalities and a latent variable, and generates the output distribution. $\Sigma$ denotes summation, and $\oplus$ denotes concatenation.}
\label{fig:decoder}
\end{figure}

\vspace{-2mm}
\paragraph{Decoding with modality-specific latent}
The attention-based decoder (\autoref{fig:decoder}) has two inputs: the set of available modalities and a latent variable,
and two key designs: (1) \textit{modality-specific latent}: the latent variable is concatenated to each modality and generates modality-specific representations independently with a shared function. (2) \textit{modality-agnostic attention}: the same attention-based modality aggregation scheme for generating full-modality representation that is robust to missing modality.

As for the multimodal encoder, the available modalities are represented as a sequence of $M^{(i)}$.
A latent code $\mathbf{\hat{z}} \sim \mathcal{N}(\mu_\phi^{(i)}, \sigma_\phi^{(i)} )$ is sampled from the posterior distribution  (or prior during testing).
The latent code is concatenated to each modality representation:
\begin{align} \label{equ14}
\mathbf J^{(i)} \!&=\! \left[ (\mathbf{\bar{x}}^{(i)}_m + \mathbf{p}^D_m) \oplus \mathbf{\hat{z}}\right]_{m \in \mathbb M^{(i)}} \in \mathbb  R^{M^{(i)} \times (d \times d_z)}
\end{align}
where $d_z$ is the dimension of the latent space, $\oplus$ is the concatenation operator and $\mathbf{P}^D = [\mathbf{p}^D_1, \ldots, \mathbf{p}^D_M] \in \mathbb  R^{M \times d}$ is a learnable position encoding.
Similarly to the second layer of the multimodal encoder, we apply multihead attention on a learnable fixed-size embedding to learn a conditional probability distribution of multimodal data.
We initialize the fixed-size embedding to be $\mathbf{P}^D$, which is also the query \textbf{Q} of the input to the decoder.

By concatenating the latent code to each modality and processing them through a separate function (we use MLPs here), we reduce the impacts among different modalities in case there is a missing modality. 
\begin{align}
\mathbf{H}_0^{(i)} \!\!&=\! \text{MLP}(\mathbf{J}^{(i)}) \\ 
\mathbf{H}_1^{(i)} \!\!&=\! \text{MHA}(\text{LN}(\mathbf{P}^D), \!\text{LN}(\mathbf{H}_{0}^{(i)}), \!\text{LN}(\mathbf{H}_{0}^{(i)})) \!+\! \mathbf{H}_{0}^{(i)}
\\
\mathbf{H}_2^{(i)} \!\!&=\! \text{MHA}(\text{LN}(\mathbf{H}_{1}^{(i)}), \!\text{LN}(\mathbf{H}_{0}^{(i)}),\! \text{LN}(\mathbf{H}_{0}^{(i)})) \!+\! \mathbf{H}_{0}^{(i)}
\end{align}
The output $\mathbf{H}_2^{(i)}$ is flattened and passed into an MLP that estimates the output distribution parameters.

\subsection{Model Learning}

Our goal is to design a model that is able to learn representations from an arbitrary number of multimodal inputs and still model the desired probability distribution $p(\mathbf{y}| \mathbb X; \Theta)$.
Using a VAE formulation, the probability distribution can be written as:
\begin{equation}
p(\mathbf{y}| \mathbb X; \Theta) = \int_\mathbf{z} p(\mathbf{y}| \mathbb{X}, \mathbf{z}; \theta) p(\mathbf{z} | \mathbb{X}; \phi) \mathrm{d}\mathbf{z}
\end{equation}
where $p(\mathbf{y}| \mathbb{X}, \mathbf{z}; \theta)$ is a complex likelihood function parameterized by $\theta$, $p(\mathbf{z} | \mathbb{X}; \phi)$ is a posterior function parameterized by $\phi$, and $\mathbf{z}$ is a latent variable.
The latent variables model both diversity and uncertainty about the future.
%\vspace{-6mm}
\vspace{-2mm}
\paragraph{Training.}
The goal is to learn the parameters of the conditional generative model.
The training process is shown on the top of \autoref{fig:overall}.
During training, the model takes as input the target $\mathbf{y}$, and the available modalities $\mathbb{X}$.
These inputs are used to compute a conditional distribution $q(\mathbf{z} | \mathbb{X}; \phi)$ from which a latent code $\mathbf{z}$ is sampled.
Since the true distribution over latent variables $\mathbf{z}$ is intractable we rely on an amortized inference network $q(\mathbf{z} | \mathbb{X}; \phi)$ that approximates it with a multivariate conditional Gaussian distribution with diagonal covariance with parameters $q(\mathbf{z} | \mathbb{X}; \phi) = \mathcal{N}(\mu_\phi, \sigma_\phi)$ where $\mu_\phi$ and $\sigma_\phi$ are functions that estimate the mean and the variance of the approximate posterior.

To prevent $\mathbf{z}$ from just copying $\mathbb{X}$, we force $q(\mathbf{z} | \mathbb{X}; \phi)$ to be close to the prior distribution $p(\mathbf{z})$ using a KL-divergence term.
We use a fixed Gaussian $\mathcal{N}(\mathbf{0}, \mathbf{I})$ as prior distribution, but it is possible to learn it.
During training, a latent variable $\mathbf{\hat{z}}$ is drawn from the approximate posterior distribution $\mathbf{\hat{z}}  \sim q(\mathbf{z} | \mathbb{X}; \phi)$.
The output prediction $\mathbf{\hat{y}}$ is then sampled from the distribution
$\mathbf{\hat{y}} \sim p(\mathbf{y}| \mathbb{X}, \mathbf{\hat{z}}; \theta)$ of our conditional generative model.

\begin{table*}[t]
\centering
\begin{tabular}{lccccccc}
\toprule
Dataset & &\multicolumn{2}{c}{\textsc{TISS}} &\multicolumn{2}{c}{\textsc{PIE}} &\multicolumn{2}{c}{\textsc{SSN}} \\
\midrule
Metric & &ADE $\downarrow$ & FDE $\downarrow$ &ADE $\downarrow$ & FDE $\downarrow$ &ADE $\downarrow$ & FDE $\downarrow$ \\
\cmidrule(lr){3-8}
\textsc{FPL} \cite{yagi2018future} & & 0.177& 0.291 & 56.66&  132.23& 0.356&  0.641 \\
\textsc{MS-LSTM} \cite{qiu2022egocentric} & &0.170&  0.259 & 43.12& 81.78 & 0.341 &  0.416  \\
\textsc{MVAE} \cite{wu2018multimodal} & &0.151&  0.242 & 29.11& 67.81 & \underline{0.317} &  0.365  \\
\textsc{CXA-Transformer} \cite{qiu2022egocentric} & & \underline{0.124}& \underline{0.205} & - & - & - & -    \\ 
\textsc{PIE} \cite{rasouli2019pie}&  & - & - & \textbf{19.50} & \textbf{45.27} & - & -   \\ 
\textsc{NavInt} \cite{zhangmultimodal}& & - & - & - & - & 0.397 & \underline{0.271}   \\ 
\midrule
Ours  & &\textbf{0.104}& \textbf{0.181} & \underline{25.13} & \underline{49.19} &\textbf{0.223}& \textbf{0.264}    \\ 
\bottomrule 
\end{tabular}
\vspace{-2mm}
\caption{\textbf{Quantitative comparison of our method and baselines on human trajectory prediction datasets} when all the modalities are used during both training and testing. Our model achieves the best or second-best performances on all datasets.
The first-best is highlighted by \textbf{bold}, and the second-best is highlighted by \underline{underline}.
% \TODO{Make this table more compact, maybe one column}
}
\label{tab:human-all-modalities}
\end{table*}

\vspace{-2mm}
\paragraph{Learning Missing Modality via Random Masking.}
The parameters of the generative model $\theta$ as well as the amortized inference network $\phi$ can be jointly optimized by maximizing the evidence lower-bound (ELBO):
\begin{align}
\mathcal{L}_{ELBO}(\mathbb{X}^{(i)}, \mathbf{y}^{(i)}) =& \mathbb{E}_{q(\mathbf{z}|\mathbb{X}^{(i)}; \phi)}[\log p(\mathbf{y}^{(i)} | \mathbb{X}^{(i)}, \mathbf{z}; \theta)] \nonumber \\
&- \text{KL}[q(\mathbf{z}|\mathbb{X}^{(i)}; \phi)||p(\mathbf{z})]
\end{align}
where KL is the Kullback-Liebler divergence between two distributions.
The KL divergence between the two Gaussian distributions is computed analytically.
We use the re-parameterization trick \citep{Kingma2014} to sample from the amortized inference network $q(\mathbf{z}|\mathbb{X}^{(i)}; \phi)$.

%\todo{add random masking paragraph here?}

To train the model to be robust to missing modalities, we randomly \textit{mask} out one of the available modalities and minimize the distance between the two distributions.
The key idea is that, for a given latent variable, the output distribution should not change too much if one of the modalities is missing.
Let $\mathbb{\check{X}}^{(i)} = \{\mathbf{x}_m^{(i)}\}_{m \in \mathbb{M}^{(i)} \setminus \{m\}}$ a subset of the available modalities where $m$ is the removed modality.
We use the KL divergence to measure the distance between the two distributions and the missing-modality loss is:
\begin{align}
&\mathcal{L}_{mis}(\mathbb{X}^{(i)}, \! \mathbb{\check{X}}^{(i)}) = \text{KL}[p(\mathbf{y} | \mathbb{X}^{(i)}\!, \mathbf{z}; \theta) ||  p(\mathbf{y} | \mathbb{\check{X}}^{(i)}\!, \mathbf{z}; \theta)]
\label{eq:loss-missing}
\end{align}
In a \textit{teacher-student} mindset, the distribution generated with all the available modalities can be seen as the teacher and the distribution generated with the missing modality can be seen as the student.
All parameters can be jointly learned in an end-to-end fashion by optimizing both losses.

\vspace{-2mm}
\paragraph{Inference.}
The goal is to generate a prediction of the future given the available modalities representing the past.
The inference (or generation) process is shown on the bottom of \autoref{fig:overall}.
First, a latent code $\mathbf{\hat{z}}$ is sampled from the prior distribution $\mathbf{\hat{z}} \sim p(\mathbf{z})$.
Then, a prediction $\mathbf{\hat{y}}$ is generated as follow: $\mathbf{\hat{y}} \sim p(\mathbf{y}| \mathbb{X}, \mathbf{\hat{z}}; \theta)$.

\section{Experiments}
\label{sec:datasets}
In this section, we evaluate our model on human trajectory prediction first, and then extend to robot manipulation tasks to show its generalizability. 
% Finally, we analyze the keys contributions of our model. 
% Then to show the robustness and generalizability of our model, we further extend our evaluation into robot manipulation task with two benchmarks.

% \paragraph{Implementation details.}
% We train our model using Adam optimizer \cite{kingma2014adam} with a batch size of 512 for 300 epochs on the human trajectory prediction tasks, and a batch size of 128 for 20 epochs on the robot manipulation tasks. 
% We train on TISS dataset with learning rate of $10^{-3}$, PIE and SNN dataset with learning rate of $10^{-4}$, robotics datasets with learning rate of $10^{-5}$. 
% Following prior works \cite{mangalam2020not, yuan2021agentformer}, we sampled 20 predictions during inference. 
% More implementation details about the network architecture are in the supplementary. 
% % More details on data preparation, network architecture, and training procedure can be found in supplementary material.  
% The model is implemented in PyTorch with Nvidia GPUs.
% \TODO{Maybe split in each sub-section}

\subsection{Human trajectory prediction}

\paragraph{Task definition.}
Let's consider an agent that is continuously moving around in a space and has been observed for a given period. 
The goal is to predict the future trajectory for the next $T$ time steps based on multimodal features $\mathbb{X}$ representing the past and the current state. 
\vspace{-2mm}
%\vspace{-3mm}
\paragraph{Datasets and metrics.}
We evaluated our model on three popular multimodal benchmarks.
\textit{TISS} \cite{qiu2022egocentric} is a human trajectory forecasting dataset collected from a egocentric viewpoint. 
It contains 12,492 unique trajectories and has three modalities: trajectory $t$, neighbour body pose $p$, and semantic map $s$. 
\textit{Pedestrian Intention Estimation (PIE)}  \cite{rasouli2019pie} is a large-scale pedestrian trajectory prediction dataset that has four modalities: trajectory $p$, grid location $g$, ego-vehicle motion $e$ and semantic map $s$. 
\textit{SFU-Store-Nav (SSN)} \cite{zhang2020sfu} is an indoor human-robot interaction dataset that contains video recordings of human behaviors and motion capture data of the human participants. The dataset has three modalities: trajectory $t$, human body pose $p$, and human head orientation $h$.
We use two standard metrics \cite{alahi2016social,amirian2019social, salzmann2020trajectron++} to evaluate the trajectory prediction performances:
Average Displacement Error (ADE) and Final Displacement Error (FDE). 
More details of datasets, metrics, preprocessing, and implementations can be found in supplementary material.

\vspace{-2mm}
\paragraph{Baseline models.} 
We compare our model with several publicly available baseline models.
Designing models to handle missing modalities for human trajectory forecasting is an under-explored area so the number of baseline models to compare with is quite limited.
\textit{Future Person Localization (FPL)} \cite{yagi2018future} is a model with a 1D convolution-deconvolution architecture that predicts future person locations %based on multimodal features from egocentric videos.
\textit{MS-LSTM} \cite{qiu2022egocentric, zhangmultimodal} is an LSTM encoder-decoder model and uses individual LSTM to encode each feature. %Encoded features are concatenated and merged using fully connected layers.
\textit{MVAE} \cite{wu2018multimodal} uses the product-of-experts to learn the joint representations from all available modalities.
\textit{CXA-Transformer} \cite{qiu2022egocentric} is a transformer-based encoder-decoder model that takes multimodal inputs with a cross-attention mechanism. %for human trajectory forecasting. 
\textit{Pedestrian Intention Estimation (PIE)} \cite{rasouli2019pie} is an RNN encoder-decoder architecture using temporal attention to learn the observation sequences.%, and a self-attention module to perform dimension reduction.
\textit{Navigational Intent Inference (NavInt)} \cite{zhangmultimodal} is a hybrid framework that uses LSTM-CNN to process multimodal features. %Features are encoded independently and concatenated for computing future prediction.
%We also report results on common baselines using only two modalities in supplementary material. 

\begin{table*}[t]
\centering
\begin{tabular}{lcccccc}
\toprule
Dataset & \multicolumn{2}{c}{\textsc{TISS}} & \multicolumn{2}{c}{\textsc{PIE}} & \multicolumn{2}{c}{\textsc{SSN}} \\
Modalities & $t,p$ & $t,s$ & $t,s,e$ & $t,s,g$ & $t,p$ & $t,h$\\
\midrule
\textsc{FPL} \cite{yagi2018future} & 0.336/0.465 & 0.341/0.481 & 77.67/150.12 & 81.79/155.35& 0.463/0.671 & 0.429/0.655 \\
\textsc{MS-LSTM} \cite{qiu2022egocentric} & 0.311/0.433 & 0.284/0.398 & 50.13/97.39 & 55.44/101.12& 0.422/0.688 & 0.418/0.631 \\
\textsc{MVAE} \cite{wu2018multimodal} & 0.277/0.412 & 0.216/0.325 & 30.51/71.28 & 48.12/88.13& \underline{0.391}/0.593 & \underline{0.344}/0.582 \\
\textsc{CXA-Transformer} \cite{qiu2022egocentric} & \underline{0.198}/\underline{0.382} & \underline{0.164}/\underline{0.291} & - & - & - & - \\
\textsc{PIE} \cite{rasouli2019pie} & - & - & \textbf{24.82}/\underline{53.74} & \textbf{27.81}/\underline{62.12} & - & - \\
\textsc{NavInt} \cite{zhangmultimodal} & - & - & - & - & 0.431/\underline{0.547} & 0.445/\underline{0.519} \\
\midrule
Ours & \textbf{0.126}/\textbf{0.242} & \textbf{0.114}/\textbf{0.187} & \underline{26.34}/\textbf{52.14} & \underline{31.96}/\textbf{58.01}& \textbf{0.337}/\textbf{0.477} & \textbf{0.311}/\textbf{0.457}\\
\bottomrule 
\end{tabular}
\vspace{-2mm}
\caption{\textbf{Quantitative ADE/FDE results for different settings of missing modalities during testing.} The first (resp. second) value is ADE (resp. FDE), and a lower value is better. $t$, $p$, $s$, $h$, $e$, and $g$ represent past trajectory, body pose, semantic map and head orientation, ego-vehicle motion, and grid location.
The first-best is highlighted by \textbf{bold}, and the second-best is highlighted by \underline{underline}.
}
\label{tab:trajectory-missing}
\end{table*}

\vspace{-2mm}
\paragraph{Results with all modalities.}
\autoref{tab:human-all-modalities} shows the results
of our models and existing models when all the modalities are available during both training and testing. 
% \autoref{tab:human-all-modalities} shows the results for human trajectory prediction datasets. \autoref{tab:robotics} shows the results on robotics datasets.
% In Tables \ref{tab:human-all-modalities}, \ref{table: PIE-baseline} and \ref{table: SNN-baseline}, we report results on TISS, PIE and SFU-Store-Nav respectively. 
Our model gets the best performance on most of the datasets, and get the second-best performance on the remaining dataset.
% On all datasets, our results shows comparable performance across all baseline methods, and demonstrate the superiority of our model.
On TISS, our model significantly improves both ADE and FDE results.
On PIE dataset, it achieves the second best performance, just after the PIE model, but this model \cite{rasouli2019pie} is complex and carefully designed for the trajectory prediction task. 
%The PIE model can only be evaluated on the PIE dataset, whereas our model is generic and work on all the datasets. 
Later we show that our model can still outperform the PIE model in the missing modality scenario without any retraining.
On SSN, our model significantly outperforms all baseline models.
% \TODO{Update writting to emphasis on the generic property of our model. The model is able to outperform specialized models.
% Q: Can the PIE model be used on other datasets or is it specific to PIE dataset? }

\vspace{-2mm}
\paragraph{Results with missing modalities.}
We evaluate the model performances when some modalities are missing during testing, similar to \cite{tsai2018learning, ma2022multimodal}.
The model has access to all modalities during training, but some modalities are missing during testing. 
For each dataset, our model is trained only once and works with different missing modalities. 
All baseline models require modification to the architecture and retraining for each setting of missing modalities. 
The results, summarized in \autoref{tab:trajectory-missing}, show our model has very good performances without retraining.
For example on TISS, we remove either neighbor body pose feature $p$ or scene semantics $s$ during testing and our model outperforms the second-best model, CXA-Transformer. 
When there is a missing modality, the ADE performance of our model only decreases between $8\%-17\%$, which is significantly better than $24\% - 37\%$ from other baseline models. 
On PIE dataset, our model has similar performances as the PIE model which is specific to this dataset, but our model significantly outperforms other baselines. 
The PIE model needs to be retrain for each setting of missing modalities, whereas our model does not need to be retrained.
On SSN, our model significantly outperforms the other models in both settings. 

\subsection{Robot manipulation}

\begin{table}[t]
\centering
\begin{tabular}{lccc}
\toprule
Modality & $i, f, p$ & $i, f$ & $i, p$ \\
\midrule
\multicolumn{4}{c}{\textsc{MuJoCo}} \\
\midrule
\textsc{LF-LSTM} \cite{liang2021multibench} & \underline{0.290} & \underline{0.583} & 0.551 \\
\textsc{MVAE} \cite{lee2020making} & 0.573 & 0.797 & 0.667 \\
\textsc{MULT} \cite{tsai2020multimodal} & 0.402 & 0.635 & \underline{0.549} \\
Ours & \textbf{0.199} & \textbf{0.287} & \textbf{0.235} \\
\midrule
\multicolumn{4}{c}{\textsc{Vision\&Touch}} \\
\midrule
\textsc{LF-LSTM} \cite{liang2021multibench} & \textbf{0.205}  & 1.794 & \underline{0.338}  \\
\textsc{MVAE} \cite{lee2020making} & 0.258 & 1.981 & 0.391\\
\textsc{MULT} \cite{tsai2020multimodal} & 0.262 & \underline{1.134} & 0.429 \\
Ours & \underline{0.237} & \textbf{0.872} &  \textbf{0.271} \\
\bottomrule
\end{tabular}
\caption{\textbf{Quantitative comparison of our method and baselines on robot manipulation datasets}. 
We report the MSE results for different set of modalities. 
$i$, $f$, and $p$ represent the image, force, and proprioception modalities.
The first-best is highlighted by \textbf{bold}, and the second-best is highlighted by \underline{underline}.
}
\label{tab:robotics}
\end{table}

\paragraph{Task definition.}
A robot arm is performing a manipulation task with an object and has been observed for a sequence of time. 
The goal is to predict either the robot's end-effector pose or the object pose in the next $T$ time steps based on the multimodal sensor inputs $\mathbb{X}$ of the past.
\vspace{-2mm}
%\vspace{-2mm}
\paragraph{Datasets and metric.}
% We evaluated our model on two multimodal datasets.
\textit{MuJoCo Push} \cite{lee2020multimodal} and \textit{Vision\&Touch} \cite{lee2020making} are both large scale multimodal datasets which contain the manipulation of simulated/real robotic arms in three modalities: image $i$, force sensor $f$, and proprioception sensors $p$. 
% More details of datasets, and preprocessing can be found in supplementary material.
Following prior works \cite{lee2020making, lee2020multimodal, liang2021multibench}, we evaluate the performance by computing the mean square error (MSE) between the prediction and the ground truth. More details on datasets, implementations, and network architectures can be found in supplementary material. 

\paragraph{Baseline models.} 
We compare our model with several publicly available baseline models.
\textit{LF-LSTM} \cite{liang2021multibench} is a late fusion LSTM model where each modalities are processed with different encoders and then fed into separate LSTMs. 
\textit{MVAE} \cite{lee2020making} encodes multimodal features using product-of-experts.
\textit{Multimodal Transformer (MULT)} \cite{tsai2019multimodal} which applies cross-attention onto different modalities, and then concatenates the resulted representations for final prediction.

% \vspace{-2mm}
% \paragraph{Implementation details.} 
% We train our model using Adam optimizer \cite{kingma2014adam} with a batch size of 128 for 20 epochs on the robot manipulation tasks. 
% We train on \textit{MuJoCo Push} with learning rate of $10^{-5}$, and \textit{Vision\&Touch} with learning rate of $5\times10^{-4}$. More details about the network architecture can be found in the supplementary. 

% \vspace{-2mm}
% \paragraph{Baseline models.} 
% We compare our model with several publicly available baseline models.
% \textit{LF-LSTM} \cite{liang2021multibench} is a late fusion LSTM model where each modalities are processed with different encoders and then feed into seperate LSTMs. 
% \textit{Sensor Fusion} \cite{lee2020making} encodes multimodal features using a variational Bayesian method.
% \textit{Multimodal Transformer (MULT)} \cite{tsai2019multimodal} which applies cross-attention onto different modalities, and then concatenates the resulted representations for final prediction.

\vspace{-2mm}
\paragraph{Results.}
\autoref{tab:robotics} shows the results
of our models and baseline models for multiple settings of missing modalities.
We use the protocol described in the human trajectory prediction section to evaluate the model performances when some modalities are missing during testing.
Our model has the best performance in five of the six settings and has the second-best on the last setting.
When some modalities are missing ($i,f$ and $i,p$ settings), the performance drop is significantly lower for our model than baseline models. 
These results demonstrate the effectiveness of our model when evaluated with missing modalities.

\subsection{Model analysis}

In this section, we analyze the importance of the multimodal fusion strategy and the robustness of the model when trained with missing modalities.
% In this section, we analyse important parameters of our
% model: 

\vspace{-2mm}
\paragraph{Fusion.} 
The fusion of a set of modalities is an important component of our model. 
\autoref{table: tiss-fusion} shows the results of different fusion choices for the TISS, and SSN datasets when replacing the multimodal encoder. 
% We show that our purposed model has better results on both datasets and has better fusion efficiency compared to other baselines. 
%\todo{maybe write more?}
Our fusion method outperforms existing fusion methods, including the cross-attention fusion method introduced in \cite{qiu2022egocentric}.
We observe that our attention-based aggregation scheme is an important component of our model because the modalities are not equally similar to each other and it is more powerful than simple aggregation or existing methods.

\vspace{-2mm}
\paragraph{Robustness.} 
%\todo{Work in progress: updating table}
To showcase the robustness of our model, we evaluate our model when it is trained with missing modalities.
%\autoref{table: tiss-train-missing} shows the results on TISS when semantic map modality is removed.
\autoref{table: tiss-train-missing} shows the results on TISS when some semantic map modality samples are missing in the training data. 
% Unlike prior work \cite{ma2021smil}, we define a ratio of modality missing as the number of samples is missing, instead of the completeness of each individual modality.
For this analysis, we define a ratio of modality missing as the number of samples is missing.
The results show that when there is missing training data, our model (1) could still maintain a good level of performances relatively to baseline models shown in \autoref{tab:human-all-modalities}, and (2) benefit greatly from having access to full modalities during training, compared to existing models that can only work with reduced training data due to missing modality.
We can see the purposed loss $\mathcal{L}_{mis}$ is important, as adding this loss increases the FDE performance by 64\% when testing with modal-complete data, and 79\% with modal-incomplete data, which validates the effectiveness to the model in multiple settings.

% % In \autoref{table: tiss-train-missing}, we show that our model not only work on missing data during inference time, but also work on incomplete data during training. We remove the semantic map modality from the TISS dataset and train our model without any modifications. 
% The results shows that when there is missing training data, our model suffers from performance drop but could still maintain a good level of performances compared to existing models shown in \autoref{tab:human-all-modalities}. 
% % The model performs badly when the model sees a modality . 
% % This is expected because the network has no knowledge of the missing modality in the training data, and it would be unrealistic to make accurate forecasting based on that. 
% We also observe that our model benefit greatly from having access to full modalities during training, and has 57\% increased in performance in terms of FDE on inference with missing modality, compared to our model that is trained with missing data. 
% We also test the effectiveness of the additional loss $\mathcal{L}_{mis}$.
% We can see removing this loss decreases significantly the performances.
% Adding this loss increases the FDE performance by 64\%, and robustness against both modal-complete and modal-incomplete data. 

\begin{figure}
\centering
\begin{subfigure}{0.3\linewidth}
    \includegraphics[width=\textwidth]{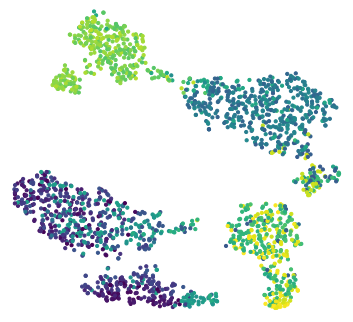}
    \caption{Modalities: $i, f$}
    \label{fig:first}
\end{subfigure}
\hfill
\begin{subfigure}{0.3\linewidth}
    \includegraphics[width=\textwidth]{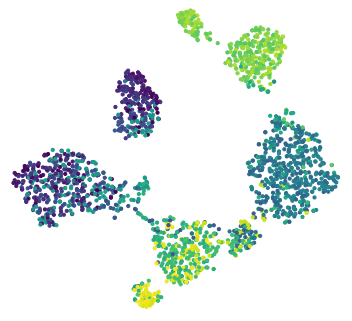}
    \caption{Modalities: $i, p$}
    \label{fig:second}
\end{subfigure}
\hfill
\begin{subfigure}{0.3\linewidth}
    \includegraphics[width=\textwidth]{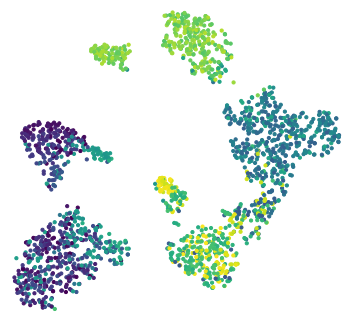}
    \caption{Modalities: $i, f, p$}
    \label{fig:third}
\end{subfigure}
\vspace{-2mm}
\caption{\textbf{t-SNE visualizations of the latent space} of our model with missing modalities on \textsc{Mujoco Push} dataset. (a) Only image $i$ and force sensor $f$ modalities. (b) Only image $i$ and proprioception sensor $p$ modalities. (c) All three modalities.}
\label{fig:viz}
\end{figure}

\vspace{-2mm}
\paragraph{Visualization of the latent space.} We visualize our robust latent distribution for \textit{MuJoCo Push} in \autoref{fig:viz}, and we qualitatively observed that the latent distributions generated from missing modalities (a)(b) are comparable to the one generated by all modalities (c). Furthermore, we can see that (b) looks more similar to (c), indicates that modality $p$ is a more important than $f$, which is also consistent with results in \autoref{tab:robotics}.

\begin{singlespace}
\begin{table}[t]
\centering
\begin{tabular}{lcccc}
\toprule
& \multicolumn{2}{c}{TISS} & \multicolumn{2}{c}{SSN} \\
Fusion & ADE $\downarrow$ & FDE $\downarrow$ & ADE $\downarrow$ & FDE $\downarrow$\\
\midrule
Linear & 0.14& 0.34 & 0.39& 0.57 \\
Average &  0.15& 0.48 & 0.41& 0.61 \\
Sum  &  0.13 & 0.31 & 0.51& 0.66\\
CXA \cite{qiu2022egocentric} & 0.12& 0.21 & 0.39 & 0.55 \\ 
\midrule
%Ours w/ Average  & 0.13& 0.24 & 0.39& 0.50 \\ 
% Ours w/ Sum  & 0.031& 0.055 & 0.27& 0.39\\ 
Ours  & \textbf{0.10}& \textbf{0.18} & \textbf{0.22} & \textbf{0.26} \\ 
\bottomrule
\end{tabular}
\vspace{-2mm}
\caption{\textbf{Quantitative comparison of fusion methods} on TISS and SSN datasets.}
\label{table: tiss-fusion}
\end{table}
\end{singlespace}

\begin{singlespace}
\begin{table}[t]
\centering
\begin{tabular}{lccc}
\toprule
Method & Training & Testing & ADE $\downarrow$ / FDE $\downarrow$ \\
\midrule
\multirow{4}{*}{Ours} & $100\%$ & $100\%$ & \textbf{0.104}/\textbf{0.181} \\
&  $100\%$ & $0\%$ & 0.126/0.232 \\
\cmidrule(lr){2-4}
&  $50\%$ & $100\%$ & 0.143/0.247\\
&  $50\%$ & $0\%$ & 0.181/0.371\\
%&  $0\%$ & $0\%$ & 0.193/0.546\\
\midrule
%\cmidrule(lr){2-4}
%CXA-T  & $50\%$ & $100\%$ & 0.217/0.465 \\
\multirow{2}{*}{MVAE \cite{lee2020making}}  & $50\%$ & $100\%$ & 0.291/0.535 \\
  & $50\%$ & $0\%$ & 0.362/0.651 \\
\midrule
%Ours w/o $\mathcal{L}_{mis}$ & 100\% & 100\% & 0.169& 0.511 \\
\multirow{2}{*}{Ours w/o $\mathcal{L}_{mis}$} & 100\% & 100\% & 0.169/0.511 \\
&  100\% & 0\% & 0.316/0.873 \\
\bottomrule 
\end{tabular}
\caption{\textbf{Ablation study} of missing training data and effectiveness of $\mathcal{L}_{mis}$ on TISS dataset. ``100\%" denotes complete modality, ``50\%" indicates 50\% of semantic map $s$ samples was removed, and ``0\%" indicates no $s$ is presented in the data. $\mathcal{L}_{mis}$ is previously defined in \autoref{eq:loss-missing}.}
\label{table: tiss-train-missing}
\end{table}
\end{singlespace}

% \begin{table}[t]
% \centering
% \begin{tabular}{lcccc}
% \toprule
% Method & Training & Testing & ADE $\downarrow$ & FDE $\downarrow$ \\
% \midrule
% \multirow{3}{*}{Ours} & Full & Full & \textbf{0.104}& \textbf{0.181} \\
% &  Full & Missing & 0.126& 0.232 \\
% % &  Missing & Full & 0.278& 0.799\\
% & Missing & Missing & 0.193 & 0.546\\ 
% \midrule
% Ours w/o $\mathcal{L}_{mis}$ & Full & Full & 0.169& 0.511 \\
% % \multirow{2}{*}{Ours w/o $\mathcal{L}_{mis}$} & Full & Full & 0.169& 0.511 \\
% % &  Full & Missing & 0.127& 0.239 \\
% \bottomrule 
% \end{tabular}
% \caption{\textbf{Ablation study} of missing training data and effectiveness of $\mathcal{L}_{mis}$ on TISS dataset. ``Full" denotes complete modality, and ``Missing" indicates the semantic map $s$ was removed. $\mathcal{L}_{mis}$ is previously defined in \autoref{eq:loss-missing}.}
% \label{table: tiss-train-missing}
% \end{table}

\section{Conclusion}

% \todo{Building efficient and accurate multimodal models will lead to better representations of the surrounding environment, that can be used to solve more complex problems.}

% \todo{Move as intro of the model section because it seems to be too detail for the introduction. 
% The overview of our proposed method is presented in Figure \ref{fig:overall}. Our Conditional Variational Autoencoder (CVAE) encodes multiple modalities of features into a single latent space, while conditioning on future groundtruth. The encoder leverages all available modalities during training and optimizes the objective function across each modality, so that the conditional likelihood can be maximized. At inference time, the model is capable of making predictions for future trajectories even when some of the modalities are missing from the input data. This is achieved by sampling from the conditional latent space and using only partial feature representations. Our proposed method thus provides a robust generative model that benefits from all input modalities during training and can handle incomplete modalities during inference.
% }

In this paper, we introduced a new attention-based model to predict the future given a set of modalities representing the current state and the past.
%that can model diverse multimodal environment, even when missing modality presented. 
% We take advantage of the adaptive nature of attention mechanism to facilitate learning across different modalities, and design both of our encoder and decoder to always learn an arbitrary number of modalities with a fixed dimension. 
% Experiments across different datasets demonstrate the robustness of our architecture. 
Different from existing multimodal models, we leverage the adaptive nature of the attention mechanism and a unique training scheme to enable learning from a set of available modalities.
We show our multimodal model is generic and works on multiple tasks: human trajectory forecasting and robot manipulation.
We demonstrate empirically that our model is effective in handling missing modalities during both training and inference.
We wish that we can further extend our work to a broader scope with different tasks, and more complicated missing modality settings.
{
    \small
    \bibliographystyle{ieeenat_fullname}
    \bibliography{main}
}

\end{document}